\pdfoutput=1

\documentclass[11pt]{article}

\usepackage[final]{acl}

\usepackage{times}
\usepackage{latexsym}

\usepackage[T1]{fontenc}

\usepackage[utf8]{inputenc}

\usepackage{microtype}

\usepackage{inconsolata}

\usepackage{graphicx}

\usepackage{dblfloatfix}
\usepackage{amsmath}
\usepackage{float}
\usepackage{booktabs}
\usepackage{listings}
\newcommand{\methodname}{Q-STRUM}

\title{\methodname\ Debate: Query-Driven Contrastive Summarization for Recommendation Comparison}

\author{George-Kirollos Saad \\
  University of Toronto \\
  Toronto, Ontario, Canada \\
  \texttt{g.saad@mail.utoronto.ca} \\\And
  Scott Sanner \\
  University of Toronto \\
  Toronto, Ontario, Canada \\
  \texttt{ssanner@mie.utoronto.ca} \\}

\begin{document}
\maketitle
\begin{abstract}

Query-driven recommendation with unknown items poses a challenge for users to understand why certain items are appropriate for their needs. Query-driven Contrastive Summarization (QCS) is a methodology designed to address this issue by leveraging language-based item descriptions to clarify contrasts between them.  However, existing state-of-the-art contrastive summarization methods such as STRUM-LLM fall short of this goal.  To overcome these limitations, we introduce \methodname\ Debate, a novel extension of STRUM-LLM that employs debate-style prompting to generate focused and contrastive summarizations of item aspects relevant to a query. Leveraging modern large language models (LLMs) as powerful tools for generating debates, \methodname\ Debate 
provides enhanced contrastive summaries. Experiments across three datasets demonstrate that \methodname\ Debate yields significant performance improvements over existing methods on key contrastive summarization criteria, thus introducing a novel and performant debate prompting methodology for QCS. 

\end{abstract}


\section{Introduction}



In query-driven recommendation settings such as hotels, restaurants, or travel, where items may be {\it a priori} unknown to users, language-based item descriptions can help users make informed choices.
However, understanding the trade-offs between choices becomes challenging given the abundance of information from objective sources, like Wikipedia or travel guides, and opinion-rich subjective sources, such as TripAdvisor and Amazon reviews~\cite{gunel2024strumllmattributedstructuredcontrastive, wen2024elaborative}. 

Fortunately, Query-driven Contrastive Summarization (QCS) offers a principled solution to these challenges by providing succinct comparative summaries of items.  
However, many traditional QCS methods often rely on complex extraction, ranking, and diversification algorithms that may fail to find clear contrasts, leaving users to sift through extensive information~\cite{qcs_summary}. 

Fortunately, the emergence of large language models (LLMs) has revolutionized 
QCS capabilities~\cite{colin2020exploring,angelidis-etal-2021-extractive,chowdhery2023palm}. 
By distilling relevant descriptive and review content into concise comparisons, state-of-the-art LLM-based contrastive summarization methods such as STRUM-LLM~\cite{gunel2024strumllmattributedstructuredcontrastive} enable users to comparatively evaluate choices via
summaries grounded in concrete data and clear comparisons that are important for decision-making~\cite{lubos2024llm, pu2006trust}.
While these LLM-driven approaches arguably improve on their pre-LLM predecessors, they often fall short of their contrastive summary potential as we show in our comparative empirical evaluation.
\begin{figure*}[ht]
    \centering
    \includegraphics[width=\linewidth]{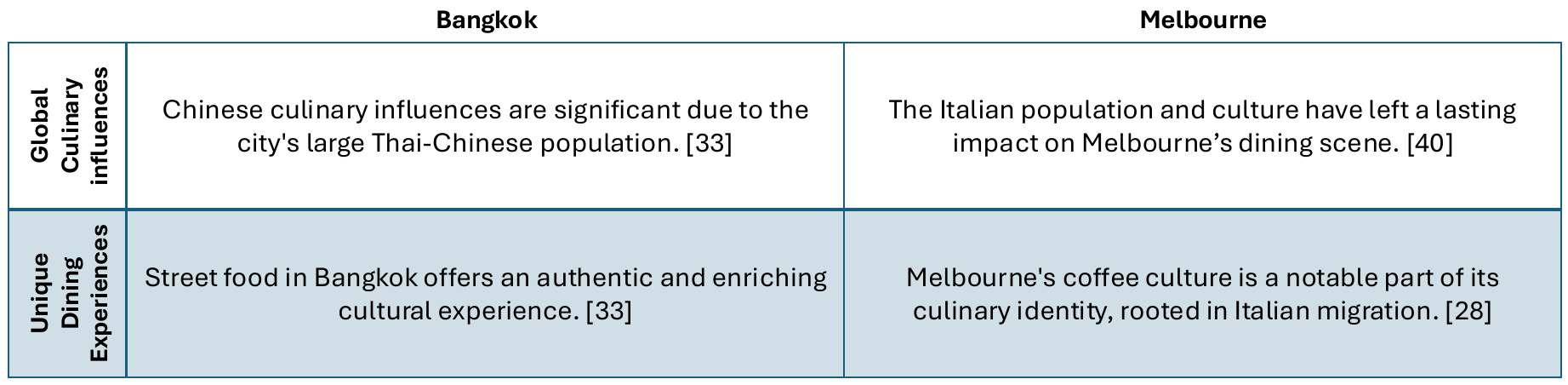}
    \caption{Example of an aspect-based contrastive summary for the query: ``culinary cities for food lovers''}
    \label{fig:good-output}
\end{figure*}

This work addresses a critical gap in QCS by leveraging LLMs to prioritize contrast while maintaining relevance and groundedness as motivated by principles of debate that are founded on discourse theory,
including the Elaboration Likelihood Model (ELM) \cite{Petty1986} and Grice's Maxims \cite{Grice2013-GRILA-2}.  Building on the state-of-the-art contrastive summarization STRUM-LLM framework~\cite{gunal-etal-2024-conversational}, we propose that aspect-based \emph{debate prompting} provides a natural framework for improved QCS that we term \methodname\ Debate.  An example output summary demonstrating this approach is shown in Figure~\ref{fig:good-output}. 

We summarize our key contributions as follows:
\begin{enumerate}
    \item We provide a novel \emph{debate prompting} mechanism to improve contrastiveness in QCS.
    \item We show the resulting \methodname\ Debate matches or outperforms base STRUM-LLM and a contrastive prompt extension on three  domains (hotels, restaurants, and travel). 
    \item We modulate debate prompt aggressiveness and evaluate its impact on summary quality.
\end{enumerate}

\section{STRUM for Contrastive Summarization}

% ...
Recommendations often involve presenting users with multiple options, requiring methods that clearly articulate how and why each option aligns with their preferences. STRUM \cite{strum} introduced a seminal and foundational approach to contrastive summarization by leveraging entailment models and hierarchical clustering to extract, merge, and contrast aspects of data. While effective for structured summarization, STRUM faced significant limitations in producing outputs that were sufficiently contrastive and aligned with user-specific queries.

\begin{figure}[!t]
    \centering
    \includegraphics[width=0.7\linewidth]{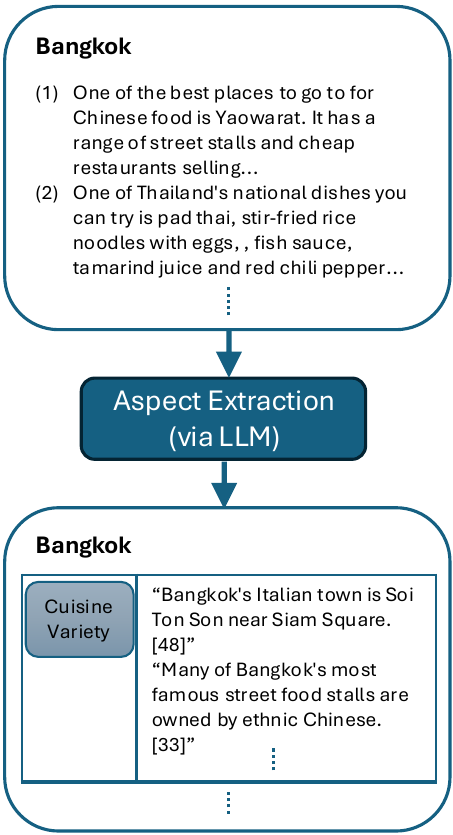}
    \caption{\textit{Aspect Extraction} stage for the query: ``culinary cities for food lovers''}
    \label{fig:aspect-extract}
\end{figure}

\begin{figure*}[!ht]
    \centering
    \includegraphics[width=\linewidth]{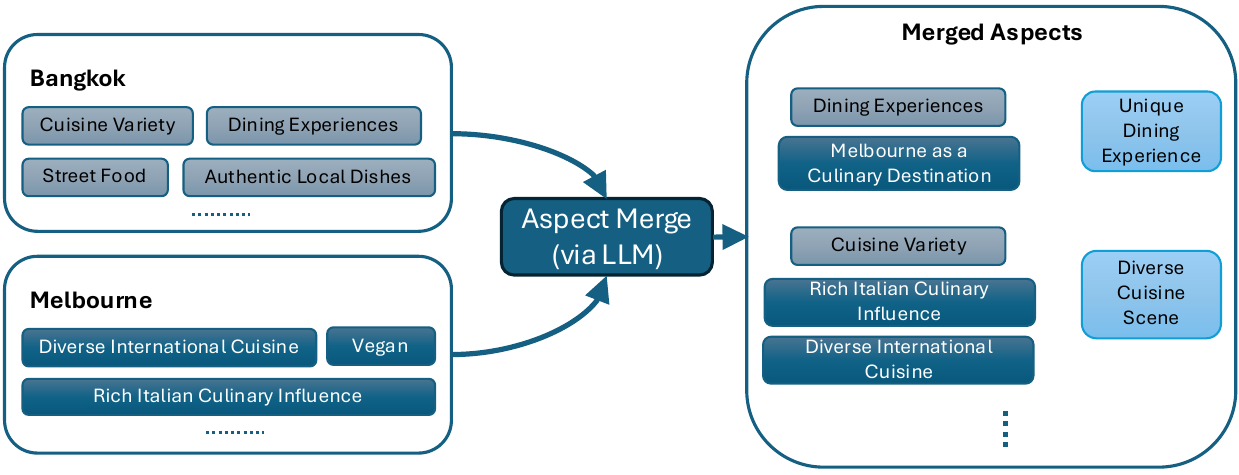}
    \caption{\textit{Aspect Merge} stage for the query: ``culinary cities for food lovers''}
    \label{fig:aspect-merge}
\end{figure*}

\begin{figure*}[!h]
    \centering
    \includegraphics[width=\linewidth]{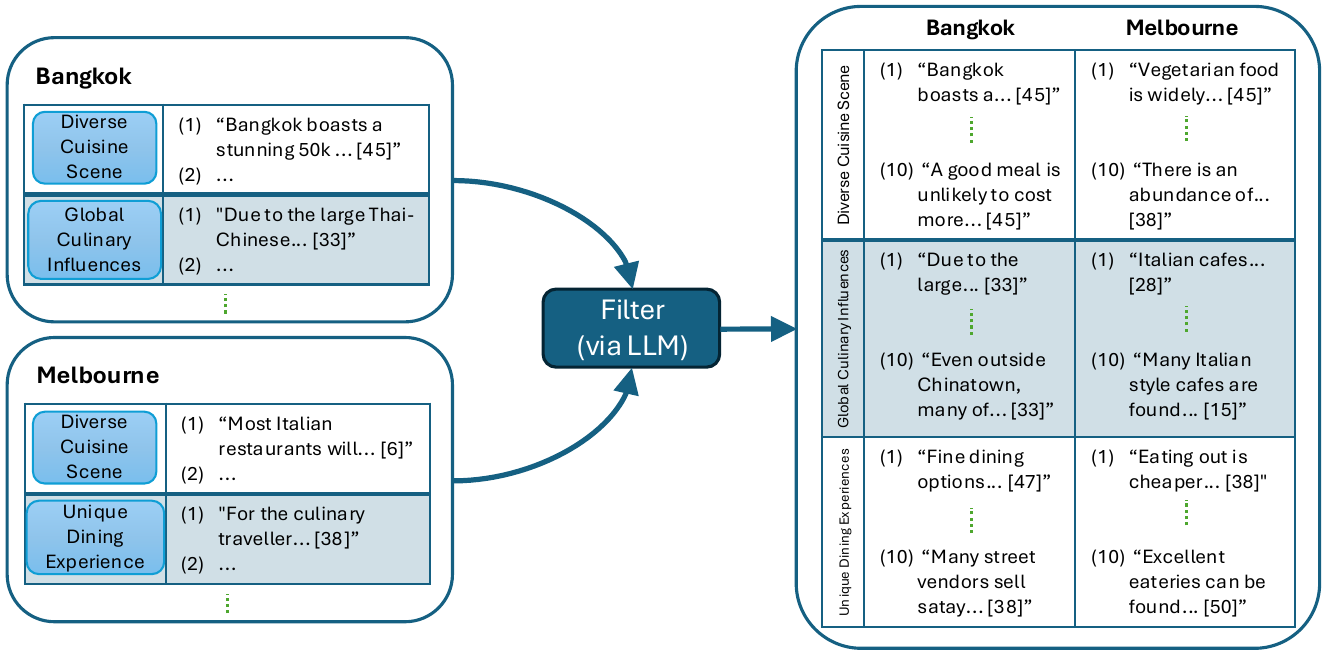}
    \caption{\textit{Filter} stage for the query: ``culinary cities for food lovers''}
    \label{fig:filter-stage}
\end{figure*}

\begin{figure*}[!h]
    \centering
    \includegraphics[width=\linewidth]{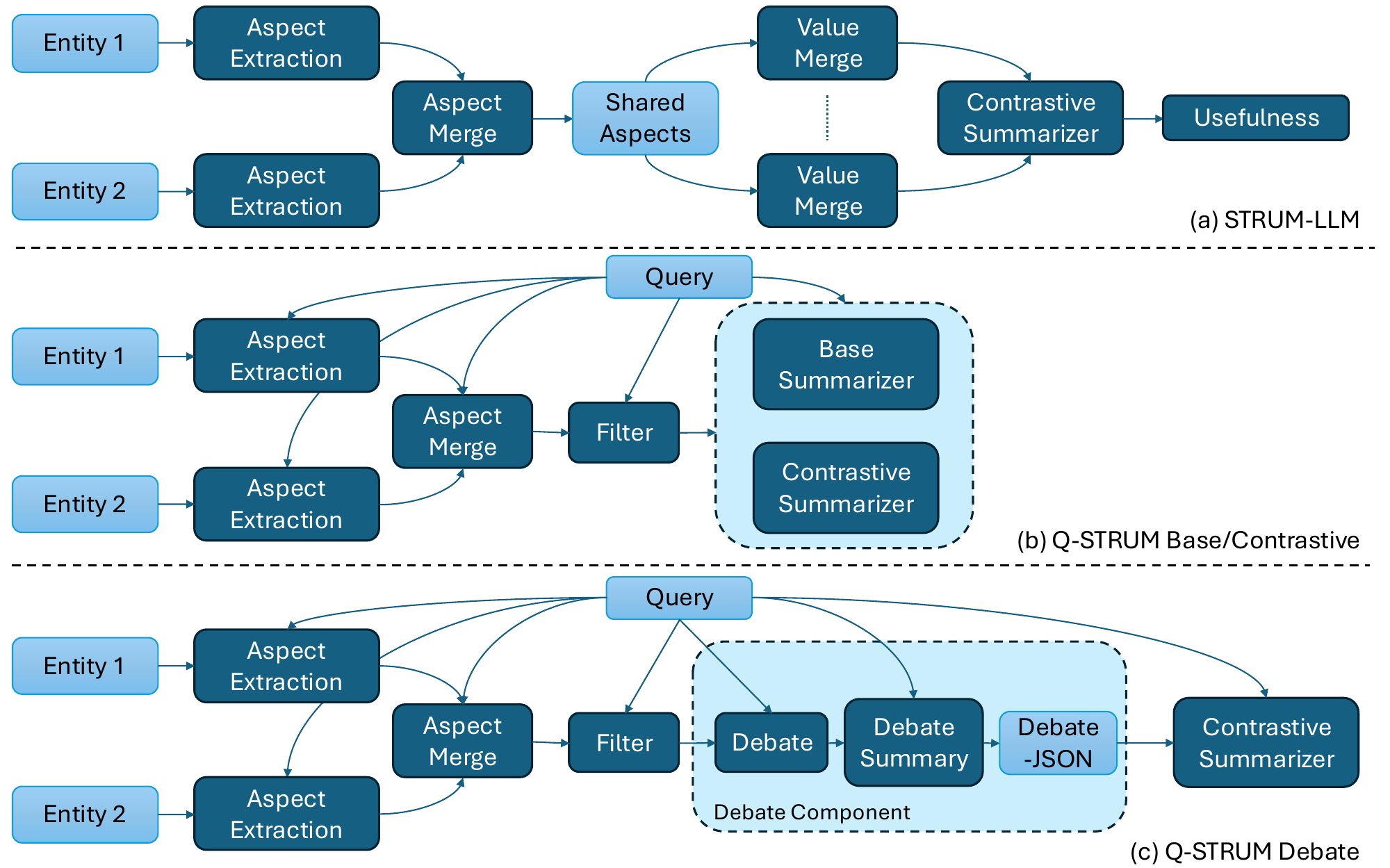}
    \caption{STRUM-LLM vs. \methodname\ Architectures}
    \label{fig:qstrum}
\end{figure*}

To address these challenges, STRUM-LLM \cite{gunel2024strumllmattributedstructuredcontrastive} integrated large language models (LLMs) to improve attribute extraction, aspect merging, and summarization. The architecture is outlined in Figure \ref{fig:qstrum}(a). STRUM-LLM employs several LLM-driven components:
\begin{itemize}
    \item \textbf{Aspect Extraction}: Identifies aspects and relevant values from source data while attributing them to their origins. An example of this stage is provided in Figure \ref{fig:aspect-extract}.
    \item \textbf{Aspect Merge}: Combines similar aspects to reduce redundancy. An example of this stage is provided in Figure \ref{fig:aspect-merge}. 
    \item \textbf{Value Merge}: Consolidates consistent values for each aspect based on majority opinion.
    \item \textbf{Contrastive Summarizer}: Highlights the most significant and contrasting aspects.
    \item \textbf{Usefulness}: Filters out less useful aspects and identifies errors.
\end{itemize}

While STRUM-LLM provides a state-of-the-art methodology for contrastive summarization, it is not query-driven as originally defined.
More critically, we also empirically show that the STRUM-LLM methodology falls short of providing strongly contrastive summaries. 
To address these gaps, we next introduce \textbf{\methodname\ Debate}, which builds upon STRUM-LLM to deliver query-driven, contrastive summarization. Central to this improvement will be the introduction of \emph{debate prompting}. 

\begin{figure*}[!t]
    \centering
    \includegraphics[width=\linewidth]{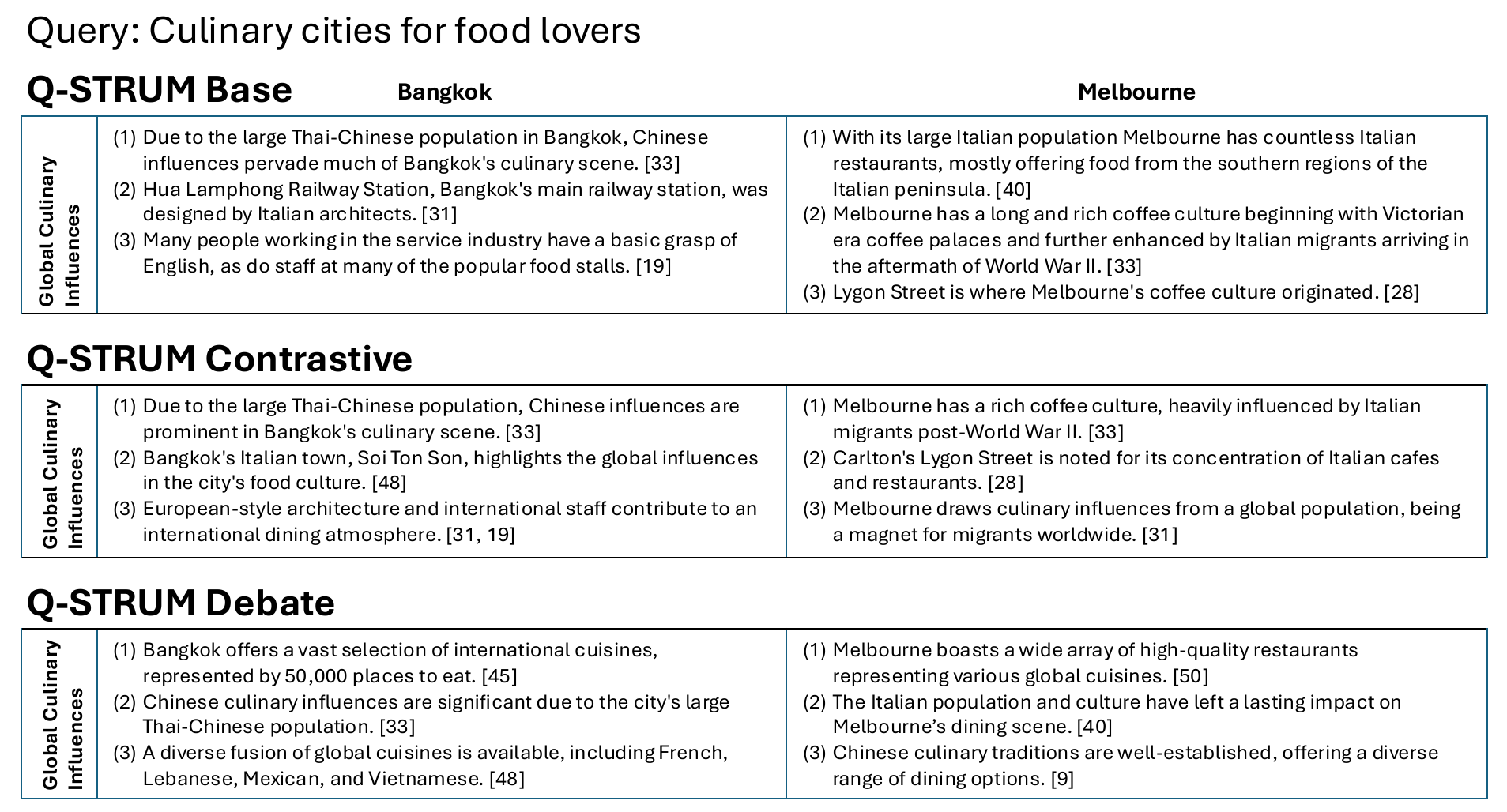}
    \caption{Comparison of Outputs of All \methodname\ Methodologies}
    \label{fig:outputs}
\end{figure*}

\section{\methodname\ for QCS}
\label{sec:q-strum}

To address the limitations of STRUM-LLM and enable a query-driven system for contrastive summarization, we propose \textbf{\methodname}. This method ensures that the query is passed through all stages of the architecture to generate highly relevant, contrastive outputs tailored to user needs.

\begin{figure*}[!t]
    \centering
    \includegraphics[width=\linewidth]{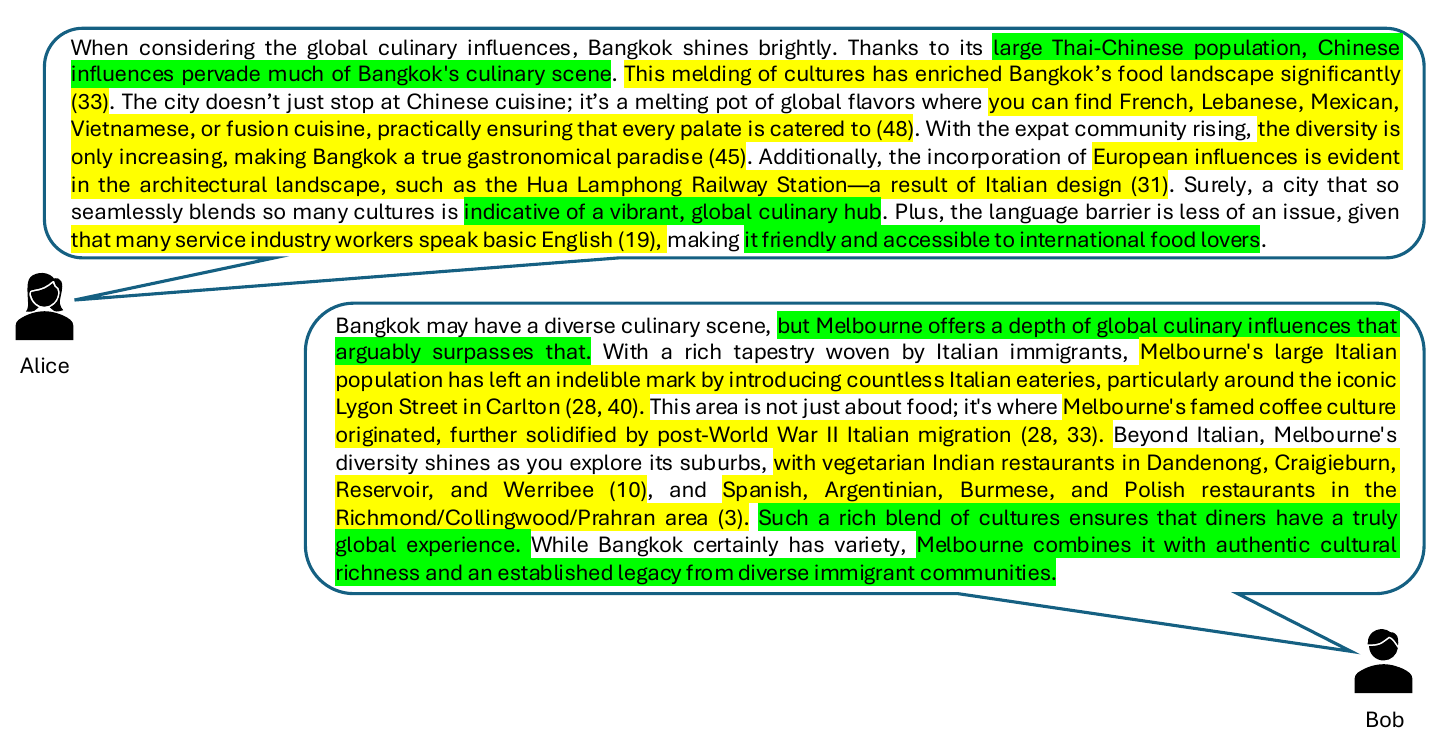}
    \caption{\textit{Debate} example for the query: ``culinary cities for food lovers''}
    \label{fig:debate-example}
\end{figure*}

\begin{figure*}[!h]
    \centering
    \includegraphics[width=\linewidth]{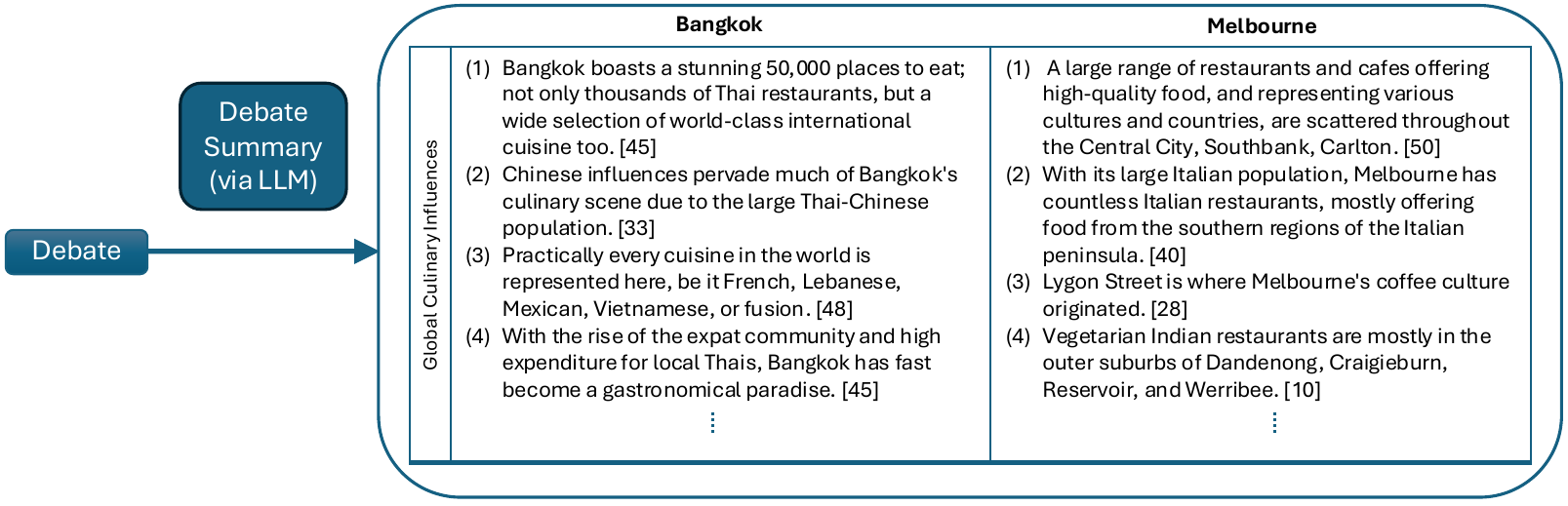}
    \caption{\textit{Debate Summary} example for the query: ``culinary cities for food lovers''}
    \label{fig:debate-json-example}
\end{figure*}

\subsection{Pipeline Overview}
\label{sec:q-strum-base}
\methodname\ employs a structured pipeline with four stages: Aspect Extraction, Aspect Merge, Filter, and Summarizer. Each stage uses the query as an anchor, ensuring alignment with user intent. The Filter stage, unique to \methodname, selects the top three aspects most relevant to the query and extracts exactly 10 concise, informative phrases for each. This reduces noise and redundancy, enabling downstream stages to focus on concise, length-controlled content. Figure \ref{fig:filter-stage} outlines an example of the Filter stage. These modifications make the Value Merge and Usefulness stages from STRUM-LLM redundant and hence allow us to remove them from the pipeline.  
Figures \ref{fig:outputs} and \ref{fig:qstrum} illustrate the output and architecture comparisons, respectively.

\begin{figure}[!ht]
    \small
    \begin{lstlisting}[caption={LLM Prompt for Debate Stage}, label={lst:debate}]
Query: {{query}}

Destination 1: {{dest1}}
{{sents1}}

Destination 2: {{dest2}}
{{sents2}}

You must simulate a debate between 2 people, Alice and Bob. 
Alice thinks that {{dest1}} is the best destination for the provided query, whereas Bob thinks {{dest2}} is for the specific aspect of: {{aspect}}. Alice and Bob should emphasize pros of their respective destinations and cons of the other destination. Make it extensive and detailed and try to mention as many sentences and points as possible. 

Perform and output a contrastive debate for each of 2 destinations for the aspect. The debate should include exact phrases from the provided sentences with sentence number citations.
    \end{lstlisting}
\end{figure}

\subsection{Contrastive Prompting}
\label{sec:q-strum-cont}

The Base Summarizer uses a monolithic prompt to produce a general summary of the extracted aspects from the Filter stage. This approach provides a simple, high-level overview of the data. The Contrastive Summarizer builds on this by explicitly instructing the LLM to \textit{``identify the most contrasting and important values.''} This simple yet effective addition produces more detailed and relevant contrastive outputs \cite{gunel2024strumllmattributedstructuredcontrastive}. Prompts for both the Base and Contrastive Summarizers are provided in Appendix \ref{sec:prompts}.

\begin{figure}[!ht]
    \small
    \begin{lstlisting}[caption={LLM Prompt for Debate Summary Stage}, label={lst:debate-json}]
Query: {{query}}
Aspect: {{aspect}}

Destination 1: {{dest1}}
{{sents1}}

Destination 2: {{dest2}}
{{sents2}}

Debate: {{debate}}

Based on the provided sentences and debate, provide a contrastive comparison for each of 2 destinations for only the listed aspect in JSON format.

Requirements are as follows:
- Do not mention Alice or Bob in the output.
- The keys should be the destination names, exactly as provided.
- The output should include summarization, backed by quotes with exact phrases from the provided sentences with sentence number citations.
- The output should be contrastive, specifically mentioning pros and cons of the destination.
- The phrasing of the output should be natural and more explanatory.
- You must include at least 5 points per aspect for each destination.


Output format:
{
 "{{dest1}}": "<extracted phrases> [sentence #]",
 "{{dest2}}": "<extracted phrases> [sentence #]"
}
    \end{lstlisting}
\end{figure}

\subsection{Debate Prompting}
\label{sec:q-strum-debate}

%
%
Debate prompting introduces a multi-stage process to address the shortcomings of monolithic summarization. Recent work demonstrates that inter-LLM debates can produce more truthful answers by leveraging structured argumentation \cite{khan2024debatingpersuasivellmsleads}. In our novel \methodname\ Debate for query-driven contrastive summarization, we simulate debate-structured disagreement between two personas to surface complementary and opposing points that are then distilled by the final summarizer.

Unlike Base and Contrastive summarizers, which rely on a single prompt, Debate prompting divides the task into distinct stages:

\begin{itemize}
    \item \textbf{Debate Stage}: The LLM simulates a structured argument between two perspectives (e.g., Alice and Bob), where each defends one entity while addressing the other's weaknesses. This ensures balanced, contrastive comparisons by emphasizing both pros and cons. An example of a debate for the query:
    \begin{quote}
        \textit{``Culinary cities for food lovers''}
    \end{quote}
    for destinations Bangkok and Melbourne is provided in Figure \ref{fig:debate-example}. Highlights include pros and cons (green) and references to source data (yellow).  The prompt in Listing~\ref{lst:debate} is called for each aspect and its respective 10 sentences output from the Filter stage (cf. Figure \ref{fig:filter-stage}).
    \item \textbf{Debate Summary Stage}: The output from the Debate Stage is summarized and then formatted into a structured JSON representation, called Debate-JSON. This step ensures that the information is well-organized and explicitly aligned with the query and extracted aspects. An example of the output of this stage for the same debate in Figure \ref{fig:debate-example}, can be found in Figure \ref{fig:debate-json-example}. The prompt used for this can be found in Listing \ref{lst:debate-json}.
    \item \textbf{Final Contrastive Summary}: The structured data from the Debate-JSON is processed using the same Contrastive Summarizer prompt from Section \ref{sec:q-strum-cont}.
\end{itemize}

This multi-stage process ensures that the outputs are not only highly contrastive but also grounded and well-structured. Figure \ref{fig:qstrum}(b) and (c) depicts the architecture of \methodname\ Debate alongside Base and Contrastive methodologies.

\begin{table*}[!ht]
\centering
\small
\begin{tabular}{@{}cccccc@{}}
\toprule
\textbf{\begin{tabular}[c]{@{}c@{}}Dataset \\\ Name\end{tabular}} &
  \textbf{\begin{tabular}[c]{@{}c@{}}Number\\ of\\ Queries\end{tabular}} &
  \textbf{\begin{tabular}[c]{@{}c@{}}Number\\ of \\ Entities\end{tabular}} &
  \textbf{\begin{tabular}[c]{@{}c@{}}Average Number of \\ Reviews / Data Snippets\\ Per Entity\end{tabular}} &
  \textbf{\begin{tabular}[c]{@{}c@{}}Average Length of\\ Review / Data Snippet\\ (in characters)\end{tabular}} &
  \textbf{Data Source} \\\midrule
TravelDest  & 50 & 774 & 163.31 & 264.53 & WikiVoyage          \\
Restaurants & 26 & 43  & 94.51  & 441.96 & TripAdvisor Reviews \\
Hotels      & 24 & 29  & 75.76  & 798.61 & TripAdvisor Reviews \\\bottomrule
\end{tabular}%
\caption{Summary of Datasets}
\label{tab:datasetssummary}
\end{table*}

\section{Datasets}
To evaluate \methodname\ Debate, we used three query-driven entity recommendation datasets with diverse and comprehensive query-entity pairs, containing detailed textual data relevant to queries.  We aimed to experiment with both objective entity descriptions as well as highly subjective review-based entity opinions.
Full data preprocessing details are provided in Appendix \ref{sec:preprocessing}.  Table \ref{tab:datasetssummary} summarizes key statistics of the following datasets: 
\begin{itemize}
    \item \textbf{TravelDest:} Objective (factual, non-review) travel destination descriptions from WikiVoyage. Example query: ``Top cities for music lovers''.
    \item \textbf{Restaurants:} Subjective restaurant reviews from TripAdvisor. Example query: ``I want a romantic restaurant with views of the city''.
    \item \textbf{Hotels:} Subjective hotel reviews from TripAdvisor. Example query: ``Find me a family-friendly hotel with enriching activities for kids''.
\end{itemize}

\section{Experimental Design}


We compare our novel \textbf{\methodname\ Debate} (Section \ref{sec:q-strum-debate}) to the query-driven STRUM-LLM extension \textbf{\methodname\ Base} (Section \ref{sec:q-strum-base}) and its  \textbf{\methodname\ Contrastive} extension (Section \ref{sec:q-strum-cont}) --- all using GPT-4o --- to address two key research questions:


    \begin{itemize}
        \item \textbf{RQ1}: Does debate-style prompting improve query-driven contrastive summaries?
        \item \textbf{RQ2}: Does the aggressiveness (or niceness) level in debate prompting impact the quality of contrastive summaries?
    \end{itemize}

\subsection{RQ1: Win Rate Evaluation Metrics}
\label{sec:evalmetrics}

We evaluate Q-STRUM and its baselines using a pairwise LLM Win Rate evaluation approach, leveraging GPT-4o and Claude-3.5-Sonnet to compare summary outputs.  
We bidirectionally tested both A vs. B and B vs. A to control for potential LLM ordering bias in winner evaluation.
This methodology aligns well with human judgments and is effective for subjective assessment tasks such as explanation evaluation \cite{liu2023gevalnlgevaluationusing,liu2024aligninghumanjudgementrole,wang2023large}. Pairwise evaluation allows for nuanced comparisons, determining a ``winner'' for each summary based on established criteria.

The Win Rate for Method A vs B is defined as:
\begin{equation}
    \text{Win Rate}_A = \frac{\text{times A wins} + 0.5 \times \text{ties}}{\text{Total Comparisons Made}} \times 100\%
\end{equation}

The evaluation focuses on four key criteria, derived from the existing literature:

\paragraph{Contrastiveness.} Summaries should effectively highlight differences, emphasizing pros and cons to help users make decisions \cite{miller2018explanationartificialintelligenceinsights,castelnovo2023evaluativeitemcontrastiveexplanationsrankings}.

\paragraph{Relevancy.} Outputs must align with the query and address user-specific needs \cite{castelnovo2023evaluativeitemcontrastiveexplanationsrankings,miller2018explanationartificialintelligenceinsights}.

\paragraph{Diversity.} Summaries should provide a variety of points without repetition, offering multiple facets of comparison \cite{Gienapp_2024,castelnovo2023evaluativeitemcontrastiveexplanationsrankings}.

\paragraph{Usefulness.} Summaries must be informative and help users in decision-making \cite{lubos2024llm,hernandez2023explaining}.

\vskip 2mm
For each query, outputs from Q-STRUM and its baselines are compared pairwise across these criteria. The LLM determines a winner for each criterion, or declares a tie, and the results are aggregated into Win Rates. To enhance evaluation quality, the LLM is prompted to justify its decisions, as reasoning summaries are shown to improve consistency \cite{zhang2022automaticchainthoughtprompting}. The prompt used for this evaluation is provided in Appendix \ref{sec:prompts}.

This approach ensures reliable, consistent, and scalable evaluation of all variations of Q-STRUM in a manner aligned with established standards for evaluating both summary and explanation quality~\cite{castelnovo2023evaluativeitemcontrastiveexplanationsrankings,miller2018explanationartificialintelligenceinsights}.





\subsection{RQ2: Aggressiveness Analysis}

This analysis examined whether varying the tone and assertiveness of debate-style prompting impacts summarization quality. Three prompt variations were tested: `nice,' `aggressive,' and the standard neutral version. The `nice' prompt instructed: ``Alice and Bob should both be nice and polite to each other.'' The `aggressive' prompt instructed: ``Alice and Bob should both be aggressive and assertive with each other.'' All other data and prompt inputs remained consistent across variations.

The standard Q-STRUM Debate was compared to Q-STRUM Debate with the modified `aggressive' and `nice' prompt versions defined above.
Evaluation followed the same pairwise Win Rate comparison over contrast, relevancy, diversity, and usefulness as in Section \ref{sec:evalmetrics}. 
Bidirectional Win Rate evaluation mitigated potential bias from aggressiveness and comparison-order interactions.

\begin{table*}[!ht]
\centering
\resizebox{\textwidth}{!}{%
\begin{tabular}{lcccccc}
\hline
           & \multicolumn{2}{c}{Restaurants}                & \multicolumn{2}{c}{Hotels}                     & \multicolumn{2}{c}{TravelDest}                 \\ \cline{2-7} 
Criterion  & Debate vs. Contrastive & Debate vs. Base       & Debate vs. Contrastive & Debate vs. Base       & Debate vs. Contrastive & Debate vs. Base       \\ \hline
Contrast   & 0.85 {[}0.78, 0.91{]}  & 0.87 {[}0.81, 0.93{]} & 0.82 {[}0.75, 0.88{]}  & 0.82 {[}0.75, 0.90{]} & 0.64 {[}0.58, 0.71{]}  & 0.78 {[}0.73, 0.84{]} \\
Relevance  & 0.57 {[}0.51, 0.63{]}  & 0.57 {[}0.51, 0.62{]} & 0.62 {[}0.55, 0.70{]}  & 0.59 {[}0.52, 0.66{]} & 0.50 {[}0.46, 0.54{]}  & 0.56 {[}0.51, 0.60{]} \\
Diversity  & 0.83 {[}0.76, 0.91{]}  & 0.84 {[}0.77, 0.91{]} & 0.80 {[}0.72, 0.88{]}  & 0.86 {[}0.79, 0.92{]} & 0.54 {[}0.48, 0.61{]}  & 0.69 {[}0.63, 0.75{]} \\
Usefulness & 0.83 {[}0.76, 0.90{]}  & 0.89 {[}0.83, 0.95{]} & 0.78 {[}0.70, 0.86{]}  & 0.84 {[}0.77, 0.91{]} & 0.61 {[}0.54, 0.68{]}  & 0.72 {[}0.66, 0.78{]} \\ \hline
\end{tabular}
}%
\caption{Pairwise LLM Win Rate (95\% CIs in $[\cdot,\cdot]$) for  Q-STRUM Debate vs. Q-STRUM Baselines (Contrastive, Base) across the Restaurants, Hotels, and TravelDest datasets using GPT-4o.}
\label{tab:debate-merged-4o}
\end{table*}

\begin{table*}[!ht]
\centering
\resizebox{\textwidth}{!}{%
\begin{tabular}{lcccccc}
\hline
           & \multicolumn{2}{c}{Restaurants}                & \multicolumn{2}{c}{Hotels}                     & \multicolumn{2}{c}{TravelDest}                 \\ \cline{2-7} 
Criterion  & Debate vs. Contrastive & Debate vs. Base       & Debate vs. Contrastive & Debate vs. Base       & Debate vs. Contrastive & Debate vs. Base       \\ \hline
Contrast   & 0.79 {[}0.71, 0.87{]}  & 0.87 {[}0.81, 0.92{]} & 0.77 {[}0.69, 0.84{]}  & 0.80 {[}0.74, 0.87{]} & 0.70 {[}0.64, 0.75{]}  & 0.75 {[}0.70, 0.79{]} \\
Relevance  & 0.63 {[}0.56, 0.69{]}  & 0.58 {[}0.53, 0.63{]} & 0.62 {[}0.56, 0.69{]}  & 0.64 {[}0.57, 0.71{]} & 0.50 {[}0.46, 0.54{]}  & 0.58 {[}0.54, 0.62{]} \\
Diversity  & 0.75 {[}0.67, 0.84{]}  & 0.84 {[}0.78, 0.91{]} & 0.77 {[}0.69, 0.84{]}  & 0.81 {[}0.74, 0.88{]} & 0.48 {[}0.42, 0.54{]}  & 0.53 {[}0.47, 0.59{]} \\
Usefulness & 0.77 {[}0.70, 0.85{]}  & 0.85 {[}0.78, 0.92{]} & 0.79 {[}0.71, 0.86{]}  & 0.84 {[}0.77, 0.91{]} & 0.54 {[}0.47, 0.60{]}  & 0.63 {[}0.57, 0.69{]} \\ \hline
\end{tabular}
}%
\caption{Pairwise LLM Win Rate (95\% CIs in $[\cdot,\cdot]$) for Q-STRUM Debate vs. Q-STRUM Baselines (Contrastive, Base) across the Restaurants, Hotels, and TravelDest datasets using Claude-3.5-Sonnet.}
\label{tab:debate-merged-claude}
\end{table*}

\begin{table*}[!h]
\centering
\resizebox{\textwidth}{!}{%
\begin{tabular}{lcccccc}
\hline
\multicolumn{1}{c}{} & \multicolumn{2}{c}{Restaurants}               & \multicolumn{2}{c}{Hotels}                    & \multicolumn{2}{c}{TravelDest}                \\
Criterion            & Aggressive            & Nice                  & Aggressive            & Nice                  & Aggressive            & Nice                  \\ \hline
Contrast             & 0.57 {[}0.48, 0.66{]} & 0.56 {[}0.46, 0.65{]} & 0.48 {[}0.33, 0.63{]} & 0.57 {[}0.49, 0.65{]} & 0.53 {[}0.47, 0.59{]} & 0.47 {[}0.41, 0.53{]} \\
Relevance            & 0.53 {[}0.48, 0.58{]} & 0.51 {[}0.45, 0.56{]} & 0.46 {[}0.37, 0.56{]} & 0.56 {[}0.49, 0.63{]} & 0.50 {[}0.46, 0.53{]} & 0.50 {[}0.47, 0.53{]} \\
Diversity            & 0.54 {[}0.45, 0.63{]} & 0.57 {[}0.48, 0.66{]} & 0.41 {[}0.26, 0.55{]} & 0.55 {[}0.46, 0.63{]} & 0.53 {[}0.47, 0.59{]} & 0.47 {[}0.41, 0.53{]} \\
Usefulness           & 0.58 {[}0.49, 0.67{]} & 0.58 {[}0.48, 0.68{]} & 0.43 {[}0.27, 0.58{]} & 0.57 {[}0.48, 0.65{]} & 0.54 {[}0.47, 0.60{]} & 0.45 {[}0.39, 0.51{]} \\ \hline
\end{tabular}
}
\caption{Pairwise LLM Win Rate (95\% CIs in $[\cdot,\cdot]$) for Q-STRUM Debate (Standard) vs. Q-STRUM Debate (Agressive and Nice) across the Restaurants, Hotels, and TravelDest datasets using GPT-4o.}
\label{tab:aggro-combined}
\end{table*}

\begin{table*}[!h]
\centering
\resizebox{\textwidth}{!}{%
\begin{tabular}{lcccccc}
\hline
\multicolumn{1}{c}{} & \multicolumn{2}{c}{Restaurants}               & \multicolumn{2}{c}{Hotels}                    & \multicolumn{2}{c}{TravelDest}                \\
Criterion            & Aggressive            & Nice                  & Aggressive            & Nice                  & Aggressive            & Nice                  \\ \hline
Contrast             & 0.55 {[}0.46, 0.65{]} & 0.58 {[}0.49, 0.67{]} & 0.50 {[}0.41, 0.59{]} & 0.58 {[}0.49, 0.67{]} & 0.53 {[}0.46, 0.59{]} & 0.54 {[}0.47, 0.61{]} \\
Relevance            & 0.52 {[}0.46, 0.57{]} & 0.55 {[}0.49, 0.60{]} & 0.47 {[}0.41, 0.53{]} & 0.55 {[}0.49, 0.62{]} & 0.50 {[}0.46, 0.53{]} & 0.50 {[}0.46, 0.53{]} \\
Diversity            & 0.55 {[}0.47, 0.64{]} & 0.49 {[}0.39, 0.59{]} & 0.47 {[}0.39, 0.55{]} & 0.59 {[}0.50, 0.68{]} & 0.58 {[}0.52, 0.64{]} & 0.53 {[}0.47, 0.59{]} \\
Usefulness           & 0.53 {[}0.43, 0.63{]} & 0.56 {[}0.46, 0.65{]} & 0.54 {[}0.45, 0.63{]} & 0.62 {[}0.52, 0.72{]} & 0.57 {[}0.51, 0.63{]} & 0.54 {[}0.48, 0.60{]} \\ \hline
\end{tabular}
}
\caption{Pairwise LLM Win Rate (95\% CIs in $[\cdot,\cdot]$) for Q-STRUM Debate (Standard) vs. Q-STRUM Debate (Aggressive and Nice) across the Restaurants, Hotels, and TravelDest datasets using Claude-3.5-Sonnet.}
\label{tab:aggro-combined-claude}
\end{table*}

\section{Experimental Results}

Below, we summarize key findings for our previous research questions.  Experiments used $\sim$11M tokens of GPT-4o (est. $\sim$200B parameters~\cite{abacha2025medecbenchmarkmedicalerror}) API calls and $\sim$3M tokens of Claude-3.5-Sonnet (175B parameters) API calls. 
 All code and data to reproduce these results are provided in a public GitHub code repository.\footnote{\url{https://github.com/D3Mlab/q-strum-debate}}  

\subsection{RQ1: Pairwise Win Rate Evaluation}

Across all datasets and criteria, Q-STRUM Debate outperformed Q-STRUM Base. For the subjective review datasets, Restaurants and Hotels, Debate achieved Win Rates mostly at or above 80\% for contrast, diversity, and usefulness, as shown in Table \ref{tab:debate-merged-4o}. However, the relevance criterion, while still favoring Debate, had lower Win Rates, at or above 57\%, however all confidence intervals were still above 50\%. 
The results for Claude-3.5-Sonnet in Table \ref{tab:debate-merged-claude} (and DeepSeek-v3 in Appendix~\ref{sec:additional-results}) are similar.
Overall, while both debate and baseline summaries are relevant, Q-STRUM Debate generally offers greater contrast, diversity, and usefulness. 

For the objective TravelDest dataset, Q-STRUM Debate demonstrated a narrower margin of superiority. The same tables highlight that Q-STRUM Debate (significantly) outperformed Q-STRUM Base in almost all criteria, but achieved mixed results against Q-STRUM Contrastive.  
This suggests that Q-STRUM Debate prompting may be more effective for the \emph{subjective, opinion-rich} TripAdvisor review data in Restaurants and Hotels than the \emph{objective, fact-oriented} WikiVoyage data of TravelDest.

\subsection{RQ2: Aggressiveness Analysis}
\label{sec:rq2}

The aggressiveness analysis compared `standard', `aggressive', and `nice' debate prompts. As shown in Tables \ref{tab:aggro-combined} and \ref{tab:aggro-combined-claude}, the standard prompt generally performed best across datasets. The aggressive prompt showed marginal improvements in specific contexts, such as the Hotels dataset, but did not demonstrate consistent advantages elsewhere. 

We conjecture that while niceness vs. aggressiveness does affect the subjective debate style and verbiage, this does not ultimately affect the core objective content being contrasted (as one can verify from the examples in Appendix \ref{sec:aggro-analysis}), hence having minimal impact on the results.

\section{Conclusion}

We introduced a novel debate-style prompting framework called Q-STRUM Debate to generate high-quality, query-driven contrastive summaries using LLMs. We demonstrated that debate-style prompting significantly outperforms baselines, particularly for subjective review datasets, by delivering more contrastive, relevant, diverse, and useful summaries. Experiments modulating debate aggressiveness showed marginal impact on results.

Future directions include multiple entity comparisons, incorporating multi-modal review content, and adapting it to diverse domains such as medical products or hiking trails. Further exploration of multi-agent debate methodology  could enhance the framework's versatility. Overall, these extensions can build on Q-STRUM's novel debate prompting methodology to improve contrastiveness in a variety of query-driven summarization applications.

\section{Limitations}

While our research introduces a robust debate model for comparative analysis, certain limitations remain that present opportunities for refinement. First, the current model is limited to pairwise comparisons of entities, which, although effective, may not fully capture the complexity of real-world decision-making scenarios where users often consider multiple options simultaneously. Additionally, our focus on textual datasets restricts the model's applicability to domains where multi-modal data (e.g., images, videos) play a critical role in user decision-making, such as in product or travel reviews.

While subjective review data enhanced the effectiveness of debates, the interpretation of conflicting subjective opinions remains an open challenge. Understanding how the model reconciles divergent viewpoints is important for improving the depth and fairness of contrastive summaries.



\section{Acknowledgments}
This work was supported by the Institute of Information \& Communications Technology Planning \& Evaluation (IITP) grant funded by the Korean Government (MSIT) (No. RS-2024-00457882, National AI Research Lab Project).

\bibliography{anthology,acl_latex}

\appendix

\section{Data Preprocessing}
\label{sec:preprocessing}

The preprocessing pipeline was designed to prepare the datasets for effective use in query-driven contrastive summarization (QCS). Key steps included:

\paragraph{TravelDest Dataset} The TravelDest dataset includes detailed WikiVoyage descriptions for 774 global destinations. For preprocessing, we employed elaborative query reformulation (EQR) \cite{wen2024elaborative} to generate rich queries. Entities were ranked for each query using dense retrieval via TAS-B embeddings \cite{hofstatter2021efficiently}. 

\paragraph{Restaurants and Hotels Datasets} For the subjective review datasets, we manually created natural language queries related to common user needs. We then scraped reviews from TripAdvisor for hotels and restaurants in Toronto. Dense retrieval was performed using OpenAI's text-embedding-3-small model\footnote{https://openai.com/index/new-embedding-models-and-api-updates/} to compute cosine similarity scores between queries and the review snippets. 

\paragraph{Snippet Extraction} For all three datasets, we extracted the top-50 relevant snippets for each entity, ensuring balanced representation. For TravelDest, these snippets were sourced from WikiVoyage, whereas for the restaurants and hotels datasets, the snippets were drawn from TripAdvisor reviews.

\paragraph{Scoring and Selection} Entity relevance was determined using the arithmetic mean of the top-50 cosine similarity scores. For each query, the top two entities were selected as candidates for comparison, and their corresponding snippets served as input for the \methodname\ pipeline.

This preprocessing approach ensures consistency, relevance, and high-quality textual inputs for evaluating \methodname's ability to generate contrastive summaries for diverse queries and data. 

\section{Aggressiveness Analysis}
\label{sec:aggro-analysis}
\begin{figure*}[!t]
    \centering
    \includegraphics[width=0.92\linewidth]{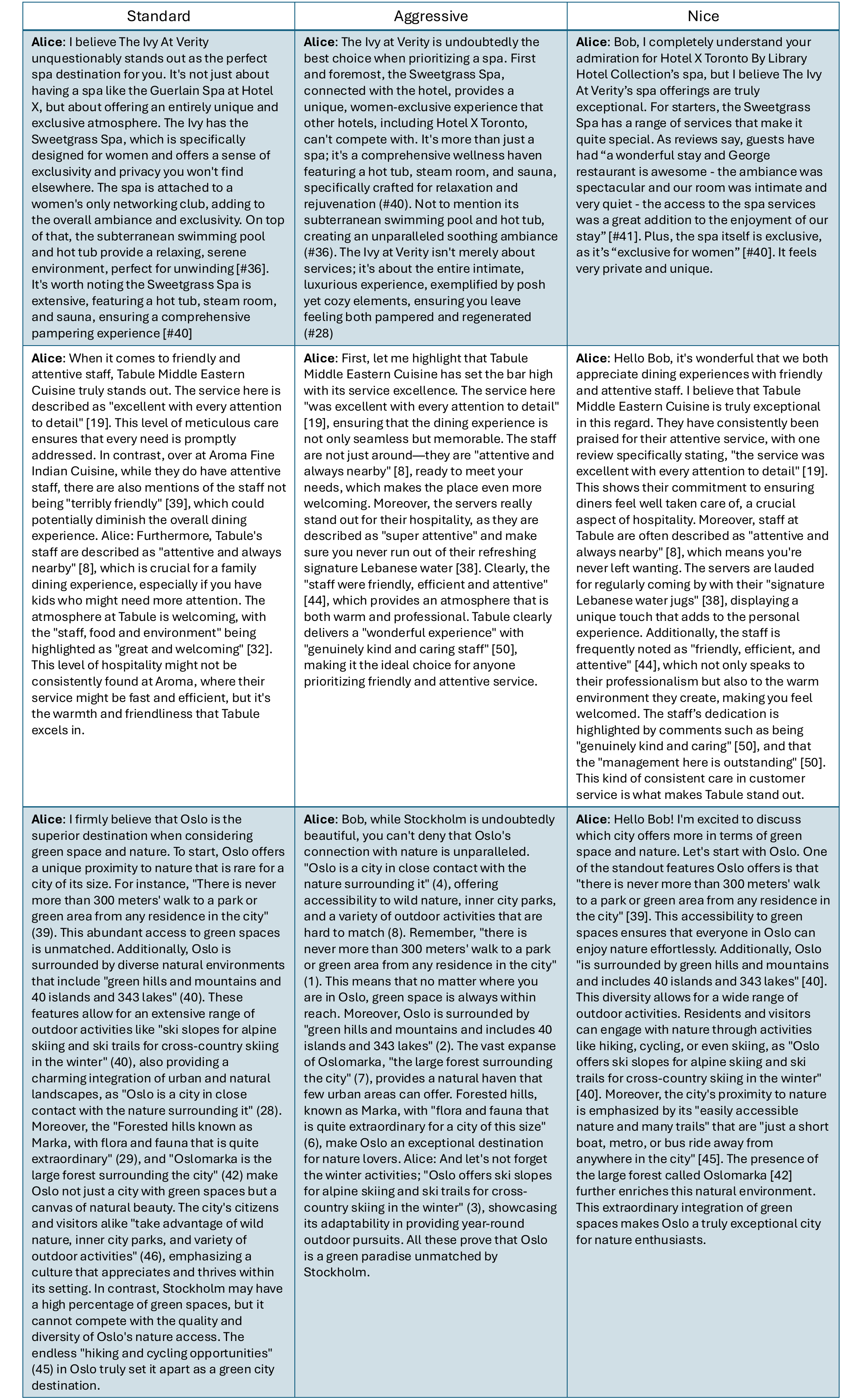}
    \caption{Debate output comparison for various aggressiveness levels}
    \label{fig:aggro-compare-debate}
\end{figure*}

Figure \ref{fig:aggro-compare-debate} shows an example comparison between `aggressive', `nice' and standard debate outputs as explored in RQ2 of Section~\ref{sec:rq2}. We see that `aggressive' prompts take a more firm stance in their wording but cover similar content as the standard prompt.  In contrast, the `nice' prompt is noticeably more collegial, but also provides more limited argumentation as evidenced by the shorter length.

\section{Q-STRUM Prompts}
\label{sec:prompts}
We provide the full prompts required to implement the Q-STRUM pipeline components as follows:
\begin{itemize}
\item Listing~\ref{lst:aspectextract}: LLM Prompt for Aspect Extraction Stage in Figure~\ref{fig:aspect-extract} and Section~\ref{sec:q-strum-base}.
\item Listing~\ref{lst:aspectmerge}: LLM Prompt for Aspect Merge Stage in Figure~\ref{fig:aspect-merge} and Section~\ref{sec:q-strum-base}.
\item Listing~\ref{lst:filter}: LLM Prompt for Filter Stage in Figure~\ref{fig:filter-stage} and Section~\ref{sec:q-strum-base}.
\item Listing~\ref{lst:contrast}: LLM Prompt for Contrastive Summarizer Stage in Figure~\ref{fig:qstrum}(b) and Section~\ref{sec:q-strum-cont}.
\item Listing~\ref{lst:llm-eval}: LLM Prompt for Pairwise Evaluation in Section~\ref{sec:evalmetrics} used for evaluation metrics.
\end{itemize}
The prompts for Q-STRUM-Debate were provided in the main paper in Listings~\ref{lst:debate} and~\ref{lst:debate-json} as discussed in Section~\ref{sec:q-strum-debate}.

\section{Additional Results}
\label{sec:additional-results}

Table \ref{tab:debate-merged-deepseek} contains the Pairwise Win Rate evaluation results using an open model, Deepseek-v3. The results are comparable to those run using GPT-4o and Claude-3.5-Sonnet.

\clearpage

\begin{figure*}[!hb]
\begin{lstlisting}[caption={LLM Prompt for Aspect Extraction Stage}, label={lst:aspectextract}]
{{destination}}
{{sentences}}

Query: {{query}}

Given the following destination and numbered texts, generate diverse and elaborative aspect phrases that describe what the user might be looking for according to the intent of the query provided and the information provided for the destination. Use the JSON format provided. 

Requirements:
- The aspect phrase must be elaborate, specific, descriptive and detailed.
- You must include the aspect and list of relevant extracted phrases for the destination for that aspect. 
- You must include a citation in a [#] format for the sentence that supports the aspect phrase from the provided sentences. Follow the same numbering as the provided sentences.
- The values must be entire, long phrases extracted exaclty from the provided sentences. 
- You must include exactly 5 aspects.
- For each aspect, you must include at least 10 extracted phrases and each extracted phrase must be highly relevant to the aspect.
- Prioritize relevancy in the extracted phrases over the number of phrases.



Output format:
{
  "<aspect>": ["extracted phrase [sentence #]", extracted phrase [sentence #]", ...],
  "<aspect>": ["extracted phrase [sentence #]", extracted phrase [sentence #]", ...],
  ...
}
    \end{lstlisting}
    \end{figure*}

\begin{figure*}[ht]
    \begin{lstlisting}[caption={LLM Prompt for Aspect Merge Stage}, label={lst:aspectmerge}]
Destination 1: {{dest1}}
Attributes 1: {{attributes1}}

Destination 2: {{dest2}}
Attributes 2: {{attributes2}}

Query: {{query}}

Merge any similar attributes from the attribute lists for each destination. Return a JSON mapping the old attribute names exactly to the new attribute names. Include the old attribute names from both destinations in the output. Ensure the new attributes are common to both destinations.

Output format:

{
  "{{dest1}}": {
    "oldAttr1": "newAttr1",
    "oldAttr2": "newAttr2",
    ...
  },
  "{{dest2}}": {
    "oldAttr3": "newAttr3",
    "oldAttr4": "newAttr4",
    ...
  }
}
    \end{lstlisting}
\end{figure*}

\begin{figure*}[ht]
    \begin{lstlisting}[caption={LLM Prompt for Filter Stage}, label={lst:filter}]
Destination 1: {{dest1}}
{{attributes1}}

Destination 2: {{dest2}}
{{attributes2}} 

Query: {{query}}

Identify the top 3 most informative attributes. For each attribute, identify exactly 10 of the most informative value pharases. You must have exactly 3 attributes per destination and exactly 10 value phrases per attribute, no exceptions. Both destinations must have the exact same 3 attributes. Follow the JSON output format provided exactly.

Output format:
{
    "{{dest1}}": {
        "<attribute1_placeholder>": ["<value phrase 1> [<citation>]", "<value phrase 2> [<citation>]", ...],
        "<attribute2_placeholder>: ["<value phrase 1> [<citation>]", "<value phrase 2> [<citation>]", ...],
        ...
    },
    "{{dest2}}": {
        "<attribute1_placeholder>": ["<value phrase 1> [<citation>]", "<value phrase 2> [<citation>]", ...],
        "<attribute2_placeholder>: ["<value phrase 1> [<citation>]", "<value phrase 2> [<citation>]",
        ...
    }
}
    \end{lstlisting}
\end{figure*}

\begin{figure*}[ht]
    \begin{lstlisting}[caption={LLM Prompt for Contrastive Summarizer Stage}, label={lst:contrast}]
Destination 1: {{dest1}}
{{attributes1}}

Destination 2: {{dest2}}
{{attributes2}} 

Query: {{query}}

Identify the most contrasting and important values and return a JSON with these attributes and their values.

Requirements are as follows:
- You must return exactly 3 attributes for each destination.
- Each attribute must have exactly 3 bullet points, summarizing both the positives and negatives of the destination for that attribute.
- Each bullet point must be relevant to the attribute and must be supported by a citation.
- The attributes should be identical for both destinations.
- Do not include meaningless attributes like null or N/A.

Output format:
{
    "{{dest1}}": {
        "<attribute1_placeholder>": ["<value phrase 1> [<citation>]", "<value phrase 2> [<citation>]", "<value phrase 3> [<citation>]"],
        "<attribute2_placeholder>: ["<value phrase 1> [<citation>]", "<value phrase 2> [<citation>]", "<value phrase 3> [<citation>]"],
        "<attribute3_placeholder>: ["<value phrase 1> [<citation>]", "<value phrase 2> [<citation>]", "<value phrase 3> [<citation>]"]
    },
    "{{dest2}}": {
        "<attribute1_placeholder>": ["<value phrase 1> [<citation>]", "<value phrase 2> [<citation>]", "<value phrase 3> [<citation>]"],
        "<attribute2_placeholder>: ["<value phrase 1> [<citation>]", "<value phrase 2> [<citation>]", "<value phrase 3> [<citation>]"],
        "<attribute3_placeholder>: ["<value phrase 1> [<citation>]", "<value phrase 2> [<citation>]", "<value phrase 3> [<citation>]"]
    }
}
    \end{lstlisting}    
\end{figure*}

\begin{figure*}[ht]
    \begin{lstlisting}[caption={LLM Prompt for Pairwise Evaluation}, label={lst:llm-eval}]
Query: {{query}}

Explanation A:
{{a}}

Explanation B:
{{b}}

Your role is to evaluate Explanation A and Explanation B as being good contrastive explanations for {{domain}} recommendation. The provided criteria should be used and you should select either "A" or "B" as the winner for each criterion or "tie" if both explanations are the same. You should provide explanations for each of your choices.

Criteria:
contrast - The summarizations should differentiate between the two {{domain}}s well, such as by including pros and cons and details, and help a user choose one {{domain}} instead of the other. 
relevancy - The summarizations provided should be relevant to each aspect and query provided.
diversity - The summarizations should provide multiple different points in support and against the {{domain}} for each aspect. Repetitive points should be penalized and a variety of different points should be rewarded. Additional context that is not repetitive should be rewarded.
usefulness - The summarizations should provide useful information and be informative for a user to make a decision between the two {{domain}}s.

Output in JSON format:
{
    "contrast": "A" or "B" or "tie",
    "contrast_explanation": <explanation>,
    "relevancy": "A" or "B" or "tie",
    "relevancy_explanation": <explanation>,
    "diversity": "A" or "B" or "tie",
    "diversity_explanation": <explanation>,
    "usefulness": "A" or "B" or "tie",
    "usefulness_explanation": <explanation>
}
    \end{lstlisting}    
\end{figure*}

\clearpage

\begin{table*}[!hb]
\centering
\resizebox{\textwidth}{!}{%
\begin{tabular}{lcccccc}
\hline
           & \multicolumn{2}{c}{Restaurants}                & \multicolumn{2}{c}{Hotels}                     & \multicolumn{2}{c}{TravelDest}                 \\ \cline{2-7} 
Criterion  & Debate vs. Contrastive & Debate vs. Base       & Debate vs. Contrastive & Debate vs. Base       & Debate vs. Contrastive & Debate vs. Base       \\ \hline
 Contrast   & 0.83 [0.76, 0.90]                  & 0.82 [0.75, 0.88]            & 0.78 [0.70, 0.86]             & 0.82 [0.75, 0.89]       & 0.57 [0.52, 0.62]                 & 0.74 [0.69, 0.79]           \\ Relevance  & 0.52 [0.49, 0.55]                  & 0.53 [0.49, 0.56]            & 0.55 [0.51, 0.59]             & 0.55 [0.50, 0.60]       & 0.50 [0.49, 0.52]                 & 0.50 [0.48, 0.51]           \\ Diversity  & 0.77 [0.70, 0.85]                  & 0.73 [0.65, 0.80]            & 0.76 [0.68, 0.84]             & 0.83 [0.77, 0.89]       & 0.51 [0.46, 0.56]                 & 0.69 [0.64, 0.74]           \\ Usefulness & 0.78 [0.71, 0.85]                  & 0.75 [0.68, 0.82]            & 0.75 [0.67, 0.83]             & 0.80 [0.74, 0.87]       & 0.50 [0.45, 0.55]                 & 0.68 [0.63, 0.73]           \\
 \hline
\end{tabular}
}%
\caption{Pairwise LLM Win Rate (95\% CIs in $[\cdot,\cdot]$) for  Q-STRUM Debate vs. Q-STRUM Baselines (Contrastive, Base) across the Restaurants, Hotels, and TravelDest datasets using Deepseek-v3.}
\label{tab:debate-merged-deepseek}
\end{table*}

\end{document}